\documentclass[11pt]{article}
\usepackage[margin=1in]{geometry}

\usepackage[T1]{fontenc}
 
\usepackage{amsmath}
\usepackage{amssymb}
 
\usepackage[numbers,sort&compress]{natbib}
 
\usepackage{amsmath,amsfonts,bm}

\def\eqref#1{equation~\ref{#1}}

\def\1{\bm{1}}

\DeclareMathAlphabet{\mathsfit}{\encodingdefault}{\sfdefault}{m}{sl}
\SetMathAlphabet{\mathsfit}{bold}{\encodingdefault}{\sfdefault}{bx}{n}

\usepackage{hyperref}
\usepackage{url}
 
\usepackage{algorithm}
\usepackage{algpseudocode}
 
\usepackage{graphicx}
\usepackage{tabularx}
\usepackage{multirow}
\usepackage{subcaption}
\usepackage{graphicx}
\usepackage{float}
\usepackage{placeins}

\usepackage{array}
\usepackage{booktabs}
\usepackage{ragged2e}
\usepackage{tabularray}
\usepackage{multirow}
 
\usepackage{tikz}
\usepackage{circuitikz}
\usepackage{pgfplots}
\pgfplotsset{compat=1.18}
 
\title{Can We Anticipate Violence? Multimodal Learning from Pre-Incident Behavioral Cues}

\author{
{\fontsize{11}{11}\selectfont
Sindhuja Penchala, Mohammed Yusuf Mujawar, Noorbakhsh Amiri Golilarz,}\\
{\fontsize{11}{11}\selectfont
Sudip Mittal, and Shahram Rahimi}\\[4pt]
{\fontsize{10}{11}\selectfont
Department of Computer Science, The University of Alabama, Tuscaloosa, AL 35487, USA}\\[3pt]
{\fontsize{9}{10}\selectfont
\{spenchala, mmujawar\}@crimson.ua.edu, \{namirigolilarz, sudip.mittal, srahimi1\}@ua.edu}
}

\date{}
 
\begin{document}
 
\maketitle

\begin{abstract}
Detecting violence after it begins is important from recognizing behavioral cues that appear immediately beforehand. This work studies short-horizon pre-incident risk recognition from multimodal video signals. We construct a binary Normal-versus-Risky setting from temporally annotated XD-Violence clips, using 443 samples with source-level separation across training, validation, and test sets. Each sample consists of a variable-length pre-incident clip, with its duration determined by the observable behavioral context preceding the incident. The incident itself is excluded from all input clips. We evaluate three complementary information sources: facial-region appearance, temporally aligned audio, and body-motion features derived from tracked keypoints. Controlled ablations are performed with Swin-Tiny, ViT-Tiny, and DeiT-Tiny to measure the contribution of each modality under the same split. Results show that combining all modalities is more effective than using any other combination alone. The best configuration, Deit-Tiny with audio, facial appearance, and motion, achieves 91.21\% accuracy, 88.96\% balanced accuracy, 93.65\% F1-score, and 96.38\% ROC-AUC on the held-out test set. These results suggest that complementary appearance, acoustic, and kinematic cues provide useful evidence for recognizing elevated pre-incident risk.
\end{abstract}

\noindent\textbf{Keywords:} Pre-incident risk recognition, multimodal learning, behavioral cues, violence anticipation, behavior analysis.

\section{Introduction}

Violent incidents in public and monitored environments can escalate rapidly, leaving only a short window for timely intervention. Automated video understanding has therefore become increasingly important for applications such as public-space surveillance, transportation hubs, campuses, healthcare facilities, and online video moderation, where continuous human monitoring is often impractical. Existing video anomaly and violence-detection methods have made substantial progress in recognizing fights, assaults, shootings, and other abnormal events, supported by large-scale benchmarks such as UCF-Crime and XD-Violence~\citep{sultani2018real, wu2020not}. Multimodal approaches further improve recognition by combining complementary visual, motion, and audio cues, particularly when the visual stream is degraded by occlusion or poor illumination~\citep{wu2020not, pang2021violence, wei2022look}. However, these systems, including online variants that operate causally, are designed to recognize violent or anomalous evidence already present in the observed sequence. For safety-critical applications, a complementary question is whether observable behavior \emph{before} an incident can distinguish emerging risk from normal activity.

Recent work has expanded violence detection beyond single visual streams by incorporating complementary audio, motion, and semantic information. Following the audio-visual XD-Violence benchmark~\citep{wu2020not}, attention-based fusion methods modeled cross-modal interactions between visual and audio features through co-attention and bilinear or additive fusion~\citep{pang2021violence, wei2022look}. Other work added caption-derived textual features to audio-visual input and strengthened multi-scale temporal modeling~\citep{na2024leveraging}. More recent methods improve cross-modal interaction through semantic feature alignment that also addresses modality asynchrony~\citep{jin2025aligning} and through gated fusion combined with multi-scale temporal modeling~\citep{ahmad2025gated}. Although these approaches improve multimodal violence recognition, they are trained and evaluated to score frames that already contain violent evidence; pre-incident frames are treated as negatives, so early responses are counted as false alarms rather than as anticipation.


Treating early responses as anticipation requires a different formulation, where the model must make a decision using only behavior observed before violent evidence appears. This problem has received less attention, although human observers have been shown to anticipate dangerous CCTV incidents above chance~\citep{troscianko2004happens}. Blunsden and Fisher studied pre-fight
recognition using spatio-temporal cuboid features and hierarchical AdaBoost~\citep{blunsden2009pre}. They showed that pre-segmented pre-fight sequences could be distinguished from normal, fight, and post-fight behavior. However, performance dropped on continuous video, where pre-fight frames were often confused with normal activity or fighting. Their continuous classifier also used a temporal window centered on the current frame, which could include observations after incident onset, and the evaluation was based on acted
scenarios. Most recent violence-detection methods are still mainly reactive and are trained
to recognize violence once it is already visible in the observed
sequence~\citep{pang2021violence,wei2022look,na2024leveraging,jin2025aligning,ahmad2025gated}.
This leaves an important gap: whether modern learned representations of human
behavior can recognize short-horizon risk using only observations captured
before an incident begins.

To address this gap, we study pre-incident risk recognition using only observations captured before annotated incident onset. The incident itself is excluded from the input, distinguishing our setting from event-present violence detection. We examine complementary cues from facial appearance, audio context, and pose-derived body motion, and evaluate them individually and in combination across multiple transformer backbones. Our objective is not to infer latent human intent, but to determine whether observable cues preceding an incident provide useful evidence of near-term risk.

\begin{itemize}
    \item We formulate pre-incident risk recognition as a setting in which models observe only behavior captured before annotated incident onset, and curate a dataset of pre-onset clips with graded risk annotations.
    \item We evaluate facial, audio, and body-motion cues individually and in combination across multiple transformer backbones to measure the contribution of each source.
    \item We show that multimodal behavioral cues observed before incident onset carry discriminative evidence of near-term risk, supporting proactive rather than reactive violence analysis.
\end{itemize}

The remainder of this paper is organized as follows.
Section~2 reviews prior work on video anomaly detection, multimodal violence
recognition, feature fusion, and event anticipation.
Section~3 presents the task formulation, multimodal representations, and
fusion framework.
Section~4 describes the dataset curation, experimental setup, ablation studies,
results, and discussions.
Finally, Section~5 concludes the paper.

\section{Related Work}
\label{sec:related_work}
In this section, we review prior work on weakly supervised anomaly detection, multimodal violence recognition, generalizable video anomaly detection, and
event anticipation.

\subsection{Weakly Supervised Video Anomaly and Violence Detection}

Weakly supervised video anomaly detection learns temporal abnormality from video-level labels without requiring dense frame annotations. Sultani et al.~\citep{sultani2018real} introduced the UCF-Crime benchmark and a multiple-instance learning formulation for anomaly detection in long, untrimmed videos. Wu et al.~\citep{wu2020not} extended this setting to audio-visual violence detection through XD-Violence, while RTFM~\citep{tian2021weakly} improved temporal anomaly localization using feature-magnitude learning. These methods provide important foundations for violence and anomaly recognition, but primarily identify abnormal evidence within the observed video.

\subsection{Multimodal Violence Detection}

Multimodal violence detection has largely been developed on XD-Violence, combining visual features with audio and, in some methods, optical flow or generated captions. Methods differ mainly in how modalities interact, including co-attention and bilinear fusion~\citep{pang2021violence}, stacked self- and co-attention units~\citep{wei2022look}, caption-augmented multi-scale temporal networks~\citep{na2024leveraging}, sparse alignment of audio and flow into the RGB feature space~\citep{jin2025aligning}, and gated fusion with multi-scale bottleneck transformers~\citep{ahmad2025gated}. Graph-based fusion operators have also been proposed for integrating heterogeneous features in video anomaly detection~\citep{ding2025learnable}. Across these methods, the emphasis is on how violent evidence is combined across modalities once it is present, rather than on what the modalities reveal beforehand.



\subsection{Video Anomaly Detection}

Recent studies have also sought to make anomaly detection transfer beyond fixed training categories, often by leveraging vision-language models. ~\citep{jiang2025local} identifies local spatial patterns through image-text alignment to generalize to novel anomalies, and LaGoVAD~\citep{liu2026language} allows anomaly definitions to be specified in natural language at inference time for open-world detection. SteerVAD~\citep{cai2026steering} instead steers anomaly-sensitive attention heads within a frozen multimodal large language model, achieving competitive performance with little training data. Because these methods ground anomalies in semantic descriptions of the event itself, they broaden what can be detected but do not address the behavior that precedes an incident.

\subsection{Action and Anomaly Anticipation}
Anticipation methods infer upcoming events from preceding observations. The Anticipative Video Transformer~\citep{girdhar2021anticipative} predicts next actions in egocentric video, and ~\citep{cao2023new} introduced video anomaly anticipation as a task, together with the semi-supervised NWPU Campus benchmark and a model that jointly detects and anticipates anomalies. More recently, PULS~\citep{samad2026latent} predicts future latent states with a video world model and reports an anticipatory advantage shortly before anomaly onset on UCF-Crime and XD-Violence. For violence specifically, ~\citep{chang2026multimodal} fuse visual, audio, and caption features and add a Siamese onset branch that models transitional dynamics around violence onset, while earlier work by ~\citep{blunsden2009pre} classified pre-fight behavior using hand-crafted features.

Our work differs from these approaches in three respects. First, in contrast to onset-aware detection, our inputs end before annotated onset and never include the incident itself. Second, whereas prior anomaly anticipation relies on generic scene representations, we focus on human-centered cues, namely facial appearance, audio, and body motion, and measure the contribution of each. Finally, rather than semi-supervised benchmarks recorded in fixed campus scenes, our clips are drawn from diverse, unconstrained XD-Violence videos spanning movies and in-the-wild footage, and are annotated with graded pre-incident risk.

\section{Methodology}
\label{sec:methodology}

\begin{figure*}[t]
    \centering
    \includegraphics[width=\textwidth]{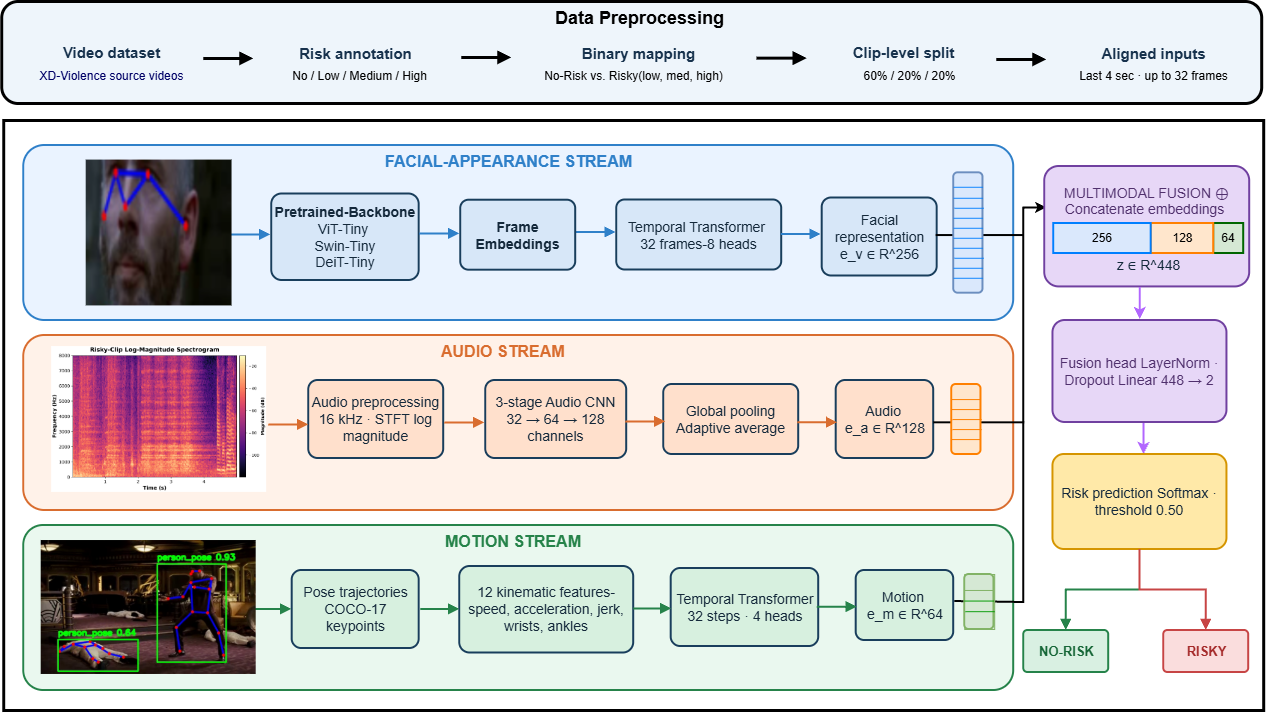}
    \caption{
    Overview of the proposed multimodal pre-incident risk classification framework.
    The preprocessing stage maps XD-Violence source videos into binary Normal and Risky
    clips and aligns facial appearance, audio, and pose-derived motion inputs.
    The facial stream uses a pretrained visual backbone followed by a temporal Transformer,
    the audio stream processes log-magnitude spectrograms using a three-stage CNN, and the
    motion stream models pose-derived kinematic features using a temporal Transformer.
    The resulting 256-dimensional visual, 128-dimensional audio, and 64-dimensional motion
    representations are concatenated and passed through a late-fusion classification head
    to predict Normal or Risky behavior.
    }
    \label{fig:architecture}
\end{figure*}

We formulate pre-incident recognition as binary clip classification using
three synchronized modalities: facial appearance, audio, and pose-derived
motion. As illustrated in Figure~\ref{fig:architecture}, each modality is encoded
independently, and the resulting clip-level representations are combined
through late feature fusion to predict either \emph{Normal} or \emph{Risky}.

\subsection{Input Construction}

For each annotated clip, we sample $T=32$ ordered frames at $8$ frames per second from an observation window ending at the clip boundary. This corresponds to a maximum duration of approximately four seconds. Shorter sequences are padded, and a validity mask prevents padded positions from contributing to temporal attention. Audio and motion features are extracted from the same interval.

\begin{equation}
\mathbf{z}_t =
\operatorname{GELU}
\left(
\mathbf{W}_v\mathbf{f}_t+\mathbf{b}_v
\right)
\in\mathbb{R}^{256}.
\end{equation}

\begin{figure}[!htbp]
    \centering

    \begin{subfigure}[t]{0.98\textwidth}
        \centering
        \includegraphics[width=\linewidth]{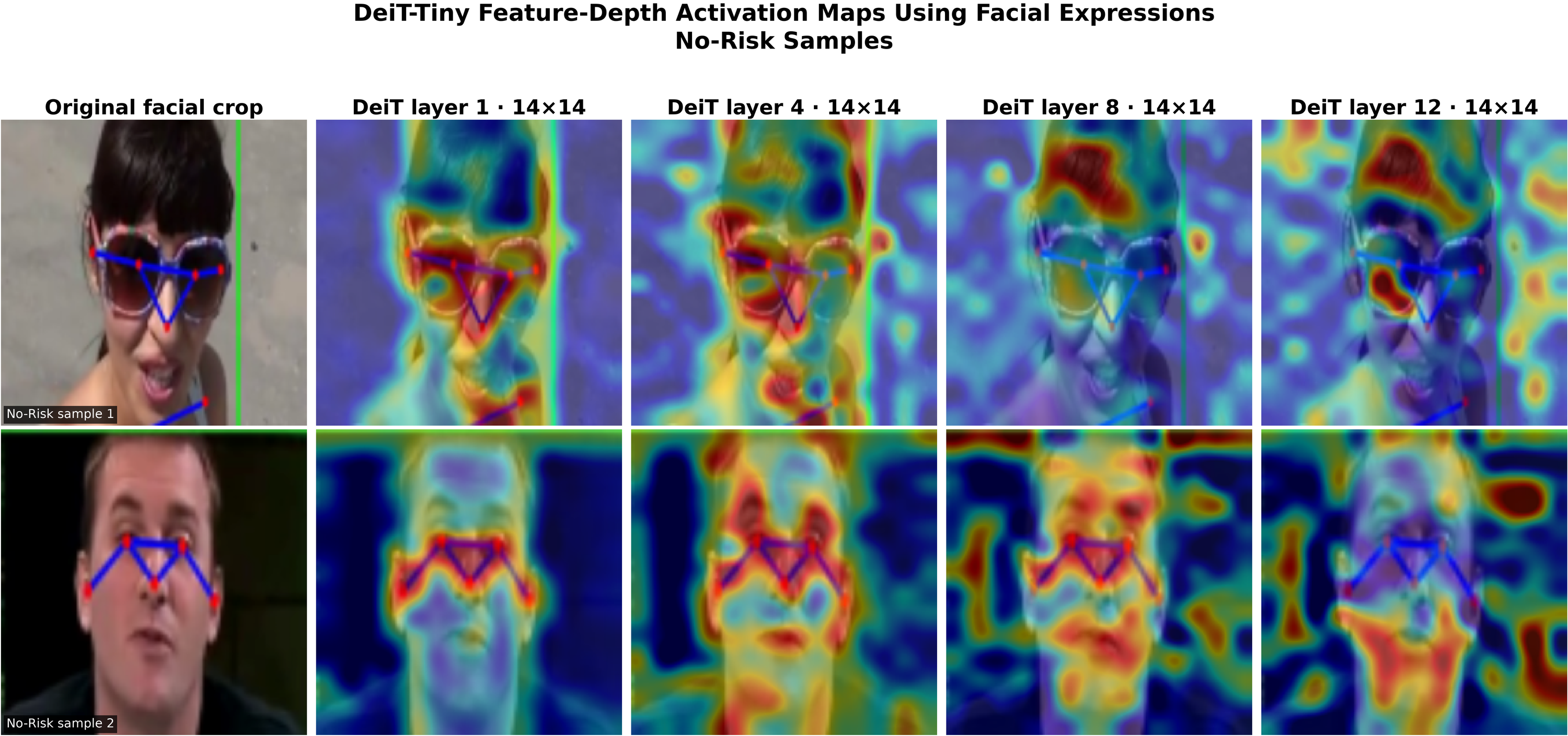}
        \caption{No-Risk samples.}
        \label{fig:deit_norisk}
    \end{subfigure}

    \vspace{0.4em}

    \begin{subfigure}[t]{0.98\textwidth}
        \centering
        \includegraphics[width=\linewidth]{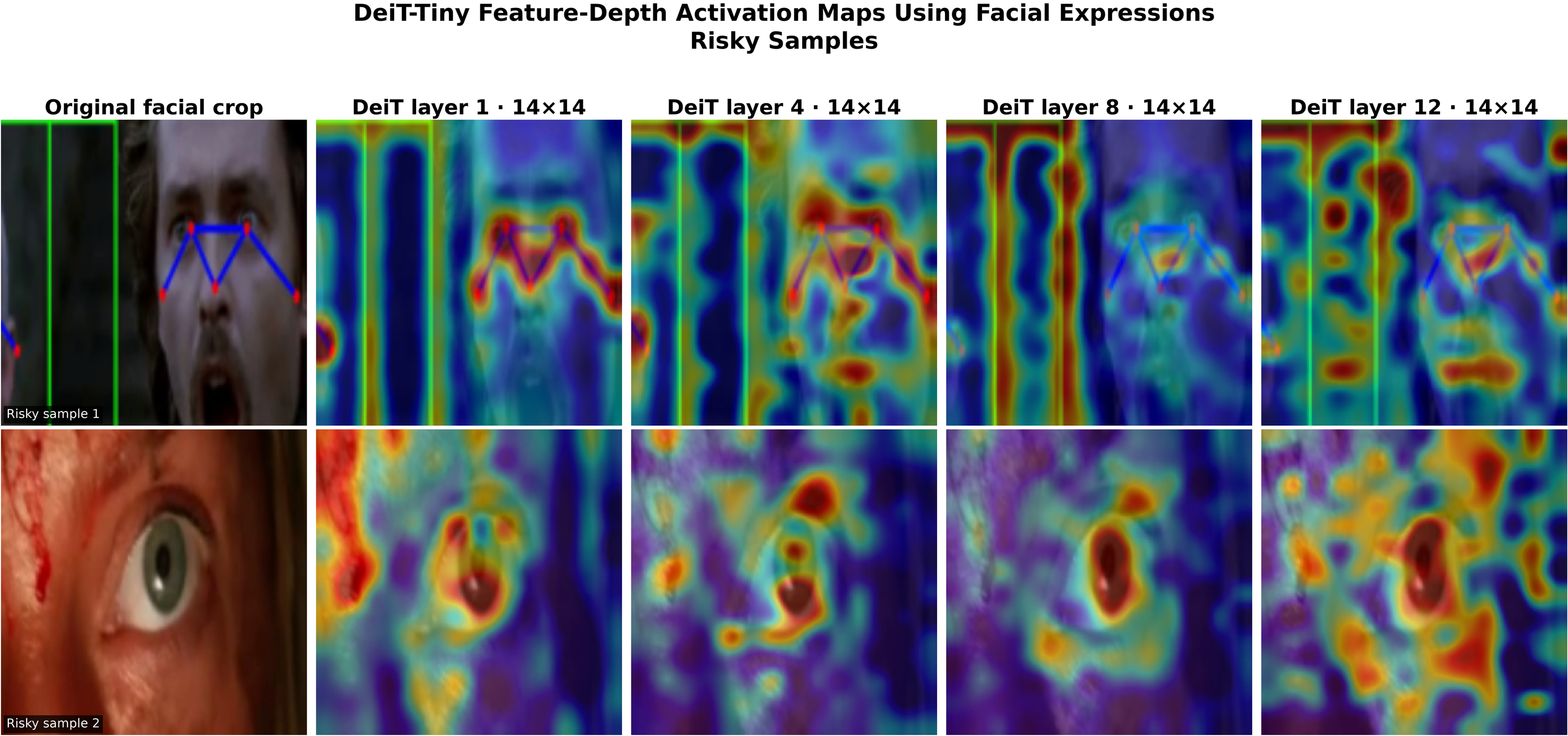}
        \caption{Risky samples.}
        \label{fig:facial_feature_maps}
    \end{subfigure}

\caption{
Feature-depth activation maps from the DeiT-Tiny facial-appearance branch for representative No-Risk and Risky samples. The first column shows the original facial crop, followed by activation maps from layers 1, 4, 8, and 12. Red and yellow regions indicate relatively strong feature responses, green and cyan indicate moderate responses, and blue and purple indicate weaker responses within each layer. These colors show activation
strength.
}
    \label{fig:deit_activation_maps}
\end{figure}
\FloatBarrier

A one-layer temporal Transformer with eight attention heads aggregates the 32 frame representations. Learned positional embeddings preserve frame order, while a classification token produces the final facial representation
$\mathbf{e}_v\in\mathbb{R}^{256}$. The pretrained visual backbone remains frozen during multimodal training to reduce overfitting. Representative layer-wise activation maps from the facial-appearance encoder are presented in Figure~\ref{fig:deit_activation_maps}.

\subsection{Facial-Appearance Stream}

Facial regions are localized using the available COCO-17 pose annotations. Visible facial landmarks are used when available; otherwise, the head region is estimated from the upper part of the person bounding box. For frames containing multiple people, the largest detected person is selected. Facial boxes are temporally smoothed, cropped with a small contextual margin, and resized to $224\times224$ pixels.

We evaluate ImageNet-pretrained Swin-Tiny, ViT-Tiny, and DeiT-Tiny backbones \cite{liu2021swin, dosovitskiy2020image, touvron2021training}. Each backbone independently encodes the facial crop at every temporal position. Its output is projected to a common 256-dimensional space:

\subsection{Audio Stream}

Audio is converted to mono, resampled to $16\,\mathrm{kHz}$, normalized, and aligned with the visual observation window. We compute a standardized log-magnitude spectrogram using a 512-point short-time Fourier transform, a 400-sample Hann window, and a 160-sample hop.

\begin{figure}[H]
    \centering
    \includegraphics[width=\textwidth]{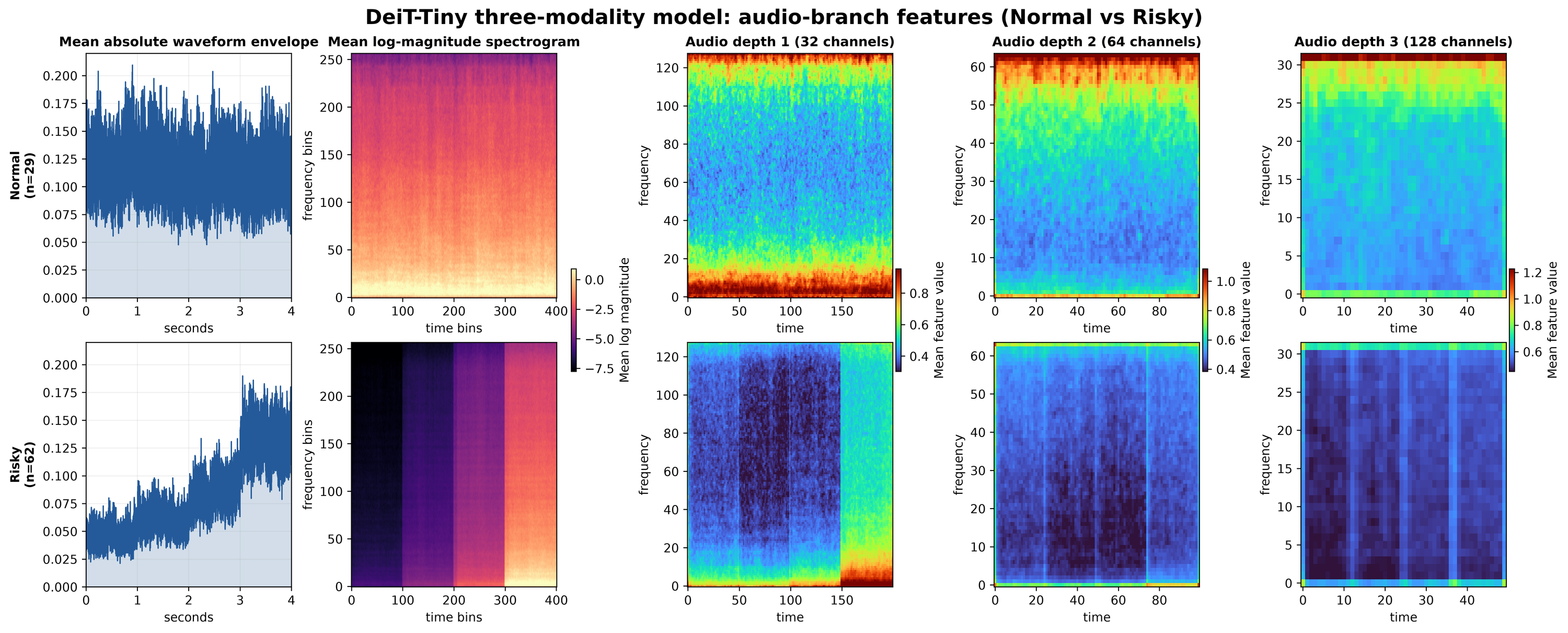}
    \caption{
    Class-averaged audio representations for Normal and Risky pre-incident
    clips using the DeiT-Tiny three-modality configuration. From left to right,
    the figure shows the mean absolute waveform, mean log-magnitude
    spectrogram, and learned audio representations at three successive network
    depths (32, 64, and 128 channels). The visualization illustrates how
    differences in the input acoustic patterns are progressively transformed
    into higher-level feature representations by the audio branch.
    }
    \label{fig:audio_feature_maps}
\end{figure}

The spectrogram is processed by a three-stage convolutional encoder with 32, 64, and 128 channels. Each stage contains convolution, batch normalization, GELU activation, and max pooling. Adaptive average pooling produces a 128-dimensional audio representation
$\mathbf{e}_a\in\mathbb{R}^{128}$. Figure~\ref{fig:audio_feature_maps} visualizes the input acoustic representations and their transformation across the three audio-CNN stages.

\begin{figure}[!htbp]
    \centering

    \begin{subfigure}[t]{0.98\textwidth}
        \centering
        \includegraphics[width=\linewidth]{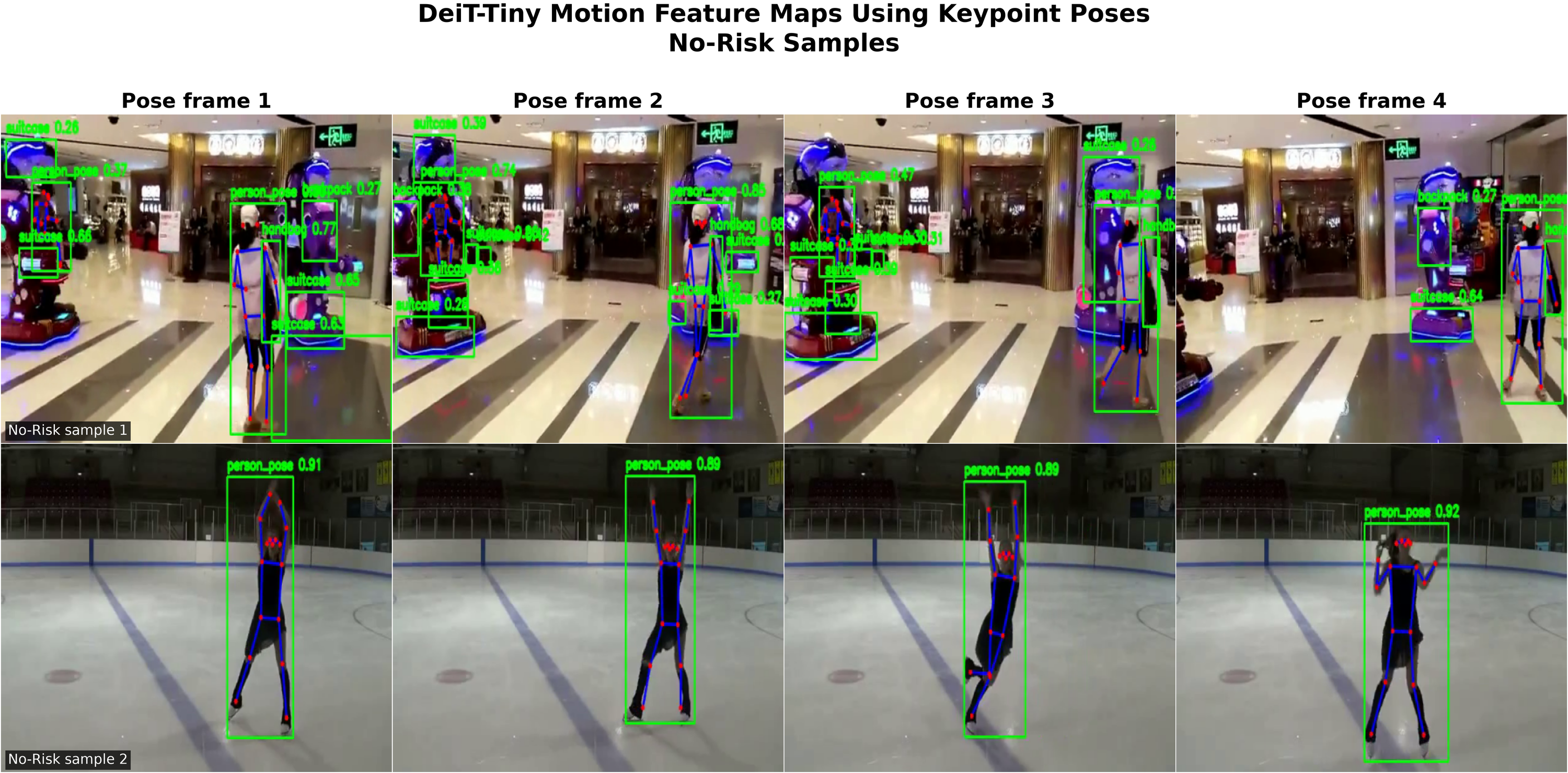}
        \caption{No-Risk samples.}
        \label{fig:motion_norisk}
    \end{subfigure}

    \vspace{0.4em}

    \begin{subfigure}[t]{0.98\textwidth}
        \centering
        \includegraphics[width=\linewidth]{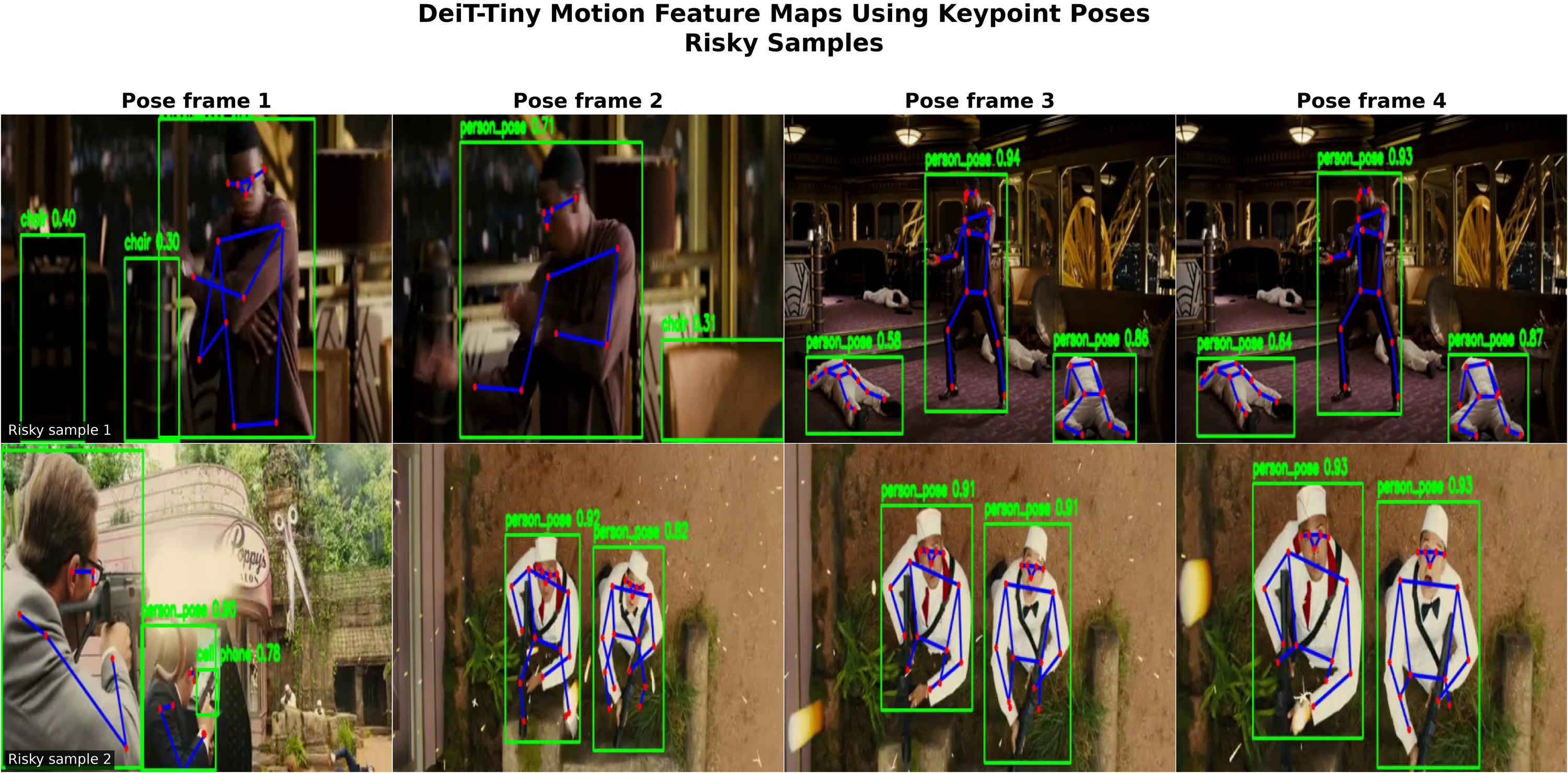}
        \caption{Risky samples.}
        \label{fig:motion_risky}
    \end{subfigure}

\caption{
Representative pose-based motion sequences for No-Risk and Risky
pre-incident clips. Four temporally ordered frames are shown for each
example. Red markers denote detected body joints, blue lines denote
skeletal connections, and green boxes indicate person detections. These
colors are visualization annotations and do not directly represent risk;
classification is based on temporal changes in the detected keypoint
positions and derived motion features.
}
    \label{fig:motion_feature_maps}
\end{figure}

\subsection{Motion Stream}

COCO-17 keypoints extracted by the YOLO pose \cite{redmon2016you} estimator are used to calculate 12 kinematic descriptors: mean and maximum speed, acceleration, and jerk; speed dispersion; mean and maximum wrist and ankle speeds; and overall keypoint-motion energy.

Motion measurements are normalized by body scale and aggregated across detected people using permutation-invariant mean and maximum statistics. After standardization using training-set statistics, the features are projected to 64 dimensions. A one-layer temporal Transformer with four attention heads produces the clip-level motion representation $\mathbf{e}_m\in\mathbb{R}^{64}$. Representative temporally ordered pose sequences used to derive these motion features are shown in Figure~\ref{fig:motion_feature_maps}.

\subsection{Multimodal Fusion and Classification}

The three modality representations are concatenated:

\begin{equation}
\mathbf{e}
=
\mathbf{e}_v
\mathbin{\Vert}
\mathbf{e}_a
\mathbin{\Vert}
\mathbf{e}_m
\in\mathbb{R}^{448},
\end{equation}

where the facial, audio, and motion streams contribute 256, 128, and 64 dimensions, respectively. Layer normalization, dropout, and a linear classifier transform the fused representation into Normal and Risky probabilities. A fixed threshold of $0.5$ is used for binary classification.

The network is optimized using class-weighted, label-smoothed cross-entropy with $\ell_2$ regularization:

\begin{equation}
\mathcal{L}
=
\mathcal{L}_{\mathrm{WCE}}
+
\lambda\mathcal{L}_{\mathrm{reg}}.
\end{equation}

\section{Experiments and Results}
\label{sec:experiments}

We evaluate the proposed framework using curated pre-incident clips from XD-Violence. The experiments examine the contribution of facial appearance, audio, and body motion, and compare different visual backbones under the same data partitions. We also provide qualitative visualizations to better understand the information captured by each modality.

\subsection{Experimental Setup}

The experimental set was constructed from the XD-Violence
dataset~\citep{wu2020not}. Source videos were manually reviewed to identify behavior occurring before a violent incident. For each Risky example, only the portion preceding the incident was retained, while the incident itself was excluded from the model input. Because the amount of observable pre-incident behavior varies across videos, the resulting clips have variable
durations.

Each clip was manually categorized as \textit{No Risk}, \textit{Low Risk}, \textit{Medium Risk}, or \textit{High Risk} using visual appearance, body motion, and audio context. For binary classification, Low, Medium, and High-Risk clips were grouped into the \textit{Risky} class, while No-Risk clips formed the \textit{Normal} class. Clips originating from the same source
video were kept within a single training, validation, or test partition to avoid information leakage. These labels represent observable pre-incident behavioral cues and are not intended to contain violence frames.

\subsection{Results and Discusion}

The complete test-set comparison across the single-, dual-, and three-modality configurations is reported in Table~\ref{tab:final_ablation_results} for all evaluated configurations. Among the tested models, the three-modality DeiT-Tiny configuration achieved the strongest overall performance, with 91.21\% accuracy, 88.96\% balanced accuracy, 92.19\% precision, 95.16\% recall, 93.65\% F1-score, and 96.38\% ROC-AUC on the held-out test set.

\begin{table*}[t]
\centering
\caption{
Final ablation results for individual and combined pre-incident modalities across different model configurations. We report Accuracy, Balanced Accuracy, Precision, Recall, F1-score, and ROC-AUC on the held-out test set. Bold values indicate the best performance obtained for each metric across all evaluated configurations.
}
\label{tab:final_ablation_results}

\scriptsize
\setlength{\tabcolsep}{4.5pt}
\renewcommand{\arraystretch}{1.18}

\resizebox{\textwidth}{!}{
\begin{tabular}{llcccccc}
\toprule
\textbf{Ablation} &
\textbf{Model} &
\textbf{Accuracy} &
\textbf{Bal. Acc.} &
\textbf{Precision} &
\textbf{Recall} &
\textbf{F1} &
\textbf{ROC-AUC} \\
\midrule

Motion only
& Temporal Transformer
& 82.42
& 76.08
& 82.86
& 93.55
& 87.88
& 86.21 \\

Audio only
& Audio CNN
& 67.03
& 73.05
& 92.11
& 56.45
& 70.00
& 78.75 \\

\midrule

Facial appearance only
& ViT-Tiny
& 87.91
& 85.62
& 90.48
& 91.94
& 91.20
& 91.94 \\

Facial appearance only
& Swin-Tiny
& 89.01
& 87.35
& 91.94
& 91.94
& 91.94
& 95.22 \\

Facial appearance only
& DeiT-Tiny
& 87.91
& 85.62
& 90.48
& 91.94
& 91.20
& 94.22 \\

\midrule

Audio + Motion
& Audio CNN + Temporal Transformer
& 78.02
& 75.61
& 85.00
& 82.26
& 83.61
& 88.71 \\

Facial appearance + Motion
& Swin-Tiny + Temporal Transformer
& 84.62
& 83.20
& 90.00
& 87.10
& 88.52
& 93.83 \\

\midrule

Audio + Facial appearance
& ViT-Tiny
& 86.81
& 83.90
& 89.06
& 91.94
& 90.48
& 93.10 \\

Audio + Facial appearance
& Swin-Tiny
& 89.01
& 86.43
& 90.62
& 93.55
& 92.06
& 96.11 \\

Audio + Facial appearance
& DeiT-Tiny
& 87.91
& 84.71
& 89.23
& 93.55
& 91.34
& 93.66 \\

\midrule

Audio + Facial appearance + Motion
& ViT-Tiny
& 86.81
& 84.82
& 90.32
& 90.32
& 90.32
& 94.33 \\

Audio + Facial appearance + Motion
& Swin-Tiny
& 90.11
& 87.24
& 90.77
& \textbf{95.16}
& 92.91
& 96.11 \\

\textbf{Audio + Facial appearance + Motion}
& \textbf{DeiT-Tiny}
& \textbf{91.21}
& \textbf{88.96}
& \textbf{92.19}
& \textbf{95.16}
& \textbf{93.65}
& \textbf{96.38} \\

\bottomrule
\end{tabular}
}
\end{table*}

\subsubsection{Ablation Studies}

We perform controlled ablation experiments to measure the contribution of three modalities: \textit{facial appearance}, \textit{audio}, and \textit{body motion}. Each modality is first evaluated independently, followed by pairwise combinations and the complete three-modality configuration. The audio branch uses the Audio CNN, while the motion branch uses the Temporal
Transformer described in Section~\ref{sec:methodology}.

For configurations containing facial appearance, we compare ViT-Tiny, Swin-Tiny, and DeiT-Tiny to examine whether multimodal performance depends on the selected visual backbone. All configurations use the same training, validation, and test partitions, allowing the effect of individual modalities
and multimodal fusion to be compared consistently.

\begin{figure*}[t]
    \centering
    \includegraphics[width=0.92\textwidth]{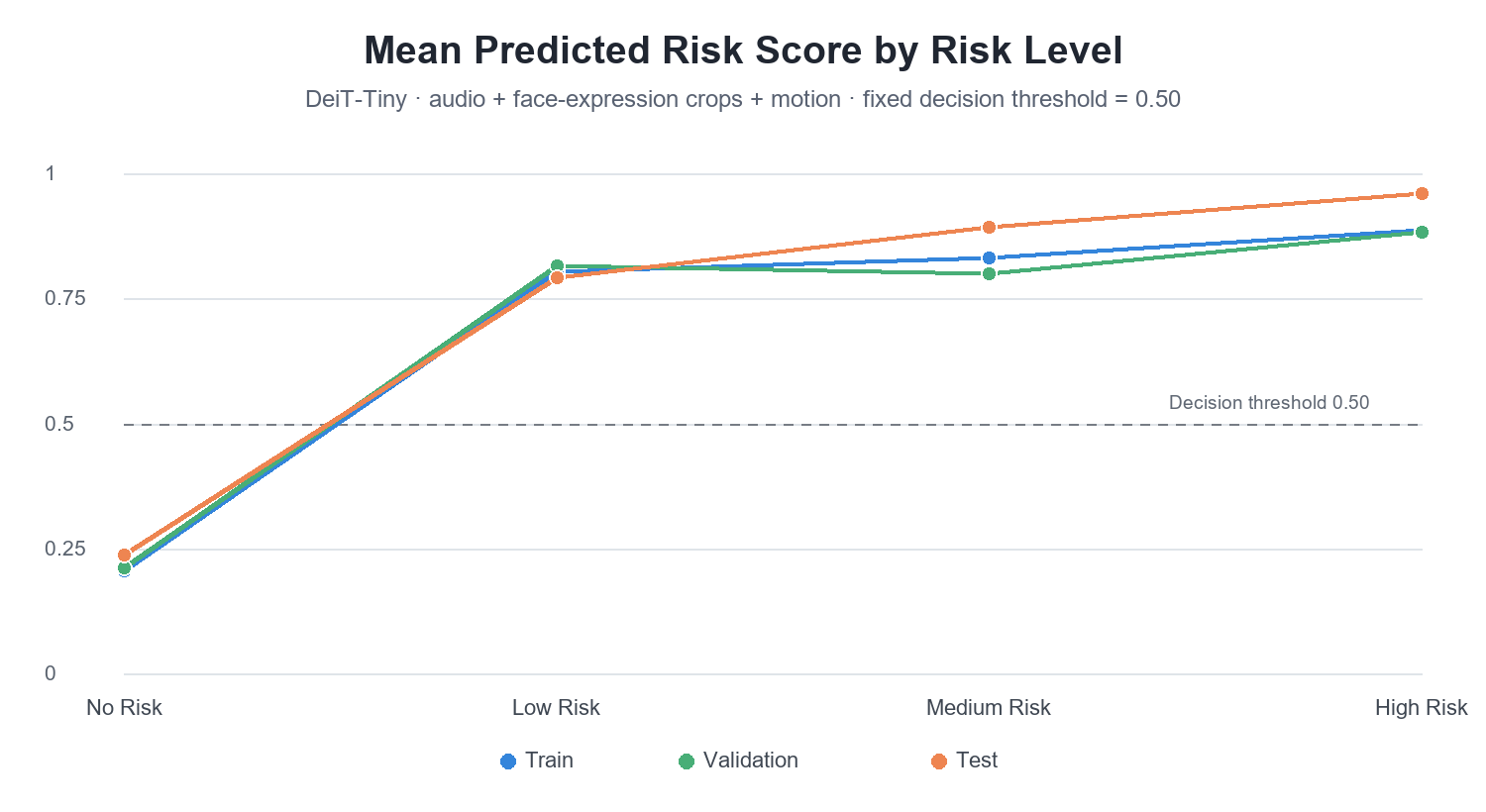}
    \caption{
    Mean predicted Risky scores across the original risk levels for the
    three-modality DeiT-Tiny model. Although the model is trained for binary
    Normal-versus-Risky classification, the average predicted risk score
    generally increases from No Risk to Low, Medium, and High Risk across the
    training, validation, and test partitions. The dashed line indicates the
    fixed decision threshold of 0.50.
    }
    \label{fig:risk_score_levels}
\end{figure*}




\subsubsection{Risk-Level Analysis}

Although the model is trained only with binary Normal and Risky labels, we
also analyze its predictions using the original No-Risk, Low-Risk,
Medium-Risk, and High-Risk annotations. As shown in
Figure~\ref{fig:risk_score_levels}, the predicted risk score is lowest for
No-Risk clips and generally increases from Low to Medium and High Risk. This
trend is consistent across the training, validation, and test sets.

This result is important because the model is never explicitly trained to
separate Low, Medium, and High Risk. Even so, its predictions follow the
progression of the original risk levels. This suggests that the model is
capturing meaningful changes in pre-incident behavior, rather than only
learning a simple Normal-versus-Risky boundary. The result also provides
evidence that useful behavioral cues can appear before the annotated incident
begins.




\subsubsection{Qualitative Analysis}
\label{sec:representation_analysis}

We examine the internal representations of the three modality branches to
understand the complementary information captured from the pre-incident
observation window. Figure~\ref{fig:deit_activation_maps} shows activation maps from successive layers of the DeiT-Tiny facial-appearance encoder for representative No-Risk
and Risky samples. Earlier layers preserve localized facial structure and
fine-grained appearance boundaries, while deeper layers produce increasingly
abstract and spatially distributed response patterns. The resulting
representations demonstrate that the visual branch captures both facial
appearance and surrounding contextual information relevant to clip-level
risk classification.

Figure~\ref{fig:audio_feature_maps} presents class-averaged audio
representations for 29 Normal and 62 Risky test clips. Normal clips exhibit
comparatively stable acoustic patterns, whereas Risky clips show stronger
temporal variation and increased acoustic activity toward the end of the
observation window. These differences remain visible across successive
convolutional stages, highlighting that the audio encoder progressively
transforms the input waveform and spectrogram into discriminative
higher-level representations.

Figure~\ref{fig:motion_feature_maps} presents temporally ordered pose
sequences used to construct the motion representation. The No-Risk examples
maintain relatively stable postures and trajectories across the observation
window. In contrast, the Risky examples exhibit more pronounced changes in
body configuration, limb movement, spatial position, and interpersonal
interaction. These temporal variations are summarized by the kinematic
descriptors and subsequently encoded by the motion Transformer.

Together, the visualizations show that the three branches capture
complementary evidence: the audio branch represents temporal acoustic
activity, the facial branch captures hierarchical appearance information,
and the motion branch models changes in body dynamics. Their combination
therefore provides a richer representation of pre-incident behaviour than
any individual modality alone.





\section{Conclusion}

This work examined whether multimodal behavioral cues observed before annotated incident onset can support short-horizon pre-incident risk recognition. By combining facial appearance, synchronized audio, and pose-derived motion, the framework captures complementary information about appearance, acoustic context, and body dynamics rather than relying on a single source of evidence. The experimental results show that multimodal integration provided better results, with the three-modality DeiT-Tiny configuration achieving the strongest overall performance of 91.21\% accuracy, 88.96\% balanced accuracy, 93.65\% F1-score, and 96.38\% ROC-AUC on the held-out test set. These findings suggest that useful discriminative information can be present in the period preceding an incident, highlighting the potential of moving violence analysis beyond event-present detection toward earlier risk recognition. In future work, we plan to expand the dataset with more diverse pre-incident scenarios and develop a more robust multimodal model capable of capturing temporal relationships across modalities and recognizing risk earlier and more reliably before incident onset.

 


\subsubsection*{Acknowledgments}
The authors acknowledge the support and resources provided by the Bioinspired Robotics, AI, Imaging and Neurocognitive Systems (BRAINS) Laboratory and Predictive Analytics and Technology Integration(PATENT) Laboratory at The University of Alabama.

\bibliography{iclr2027_conference}
\bibliographystyle{iclr2027_conference}

\clearpage

\appendix
\section{appendix}

\begin{figure*}[t]
    \centering
    \includegraphics[width=0.88\textwidth]{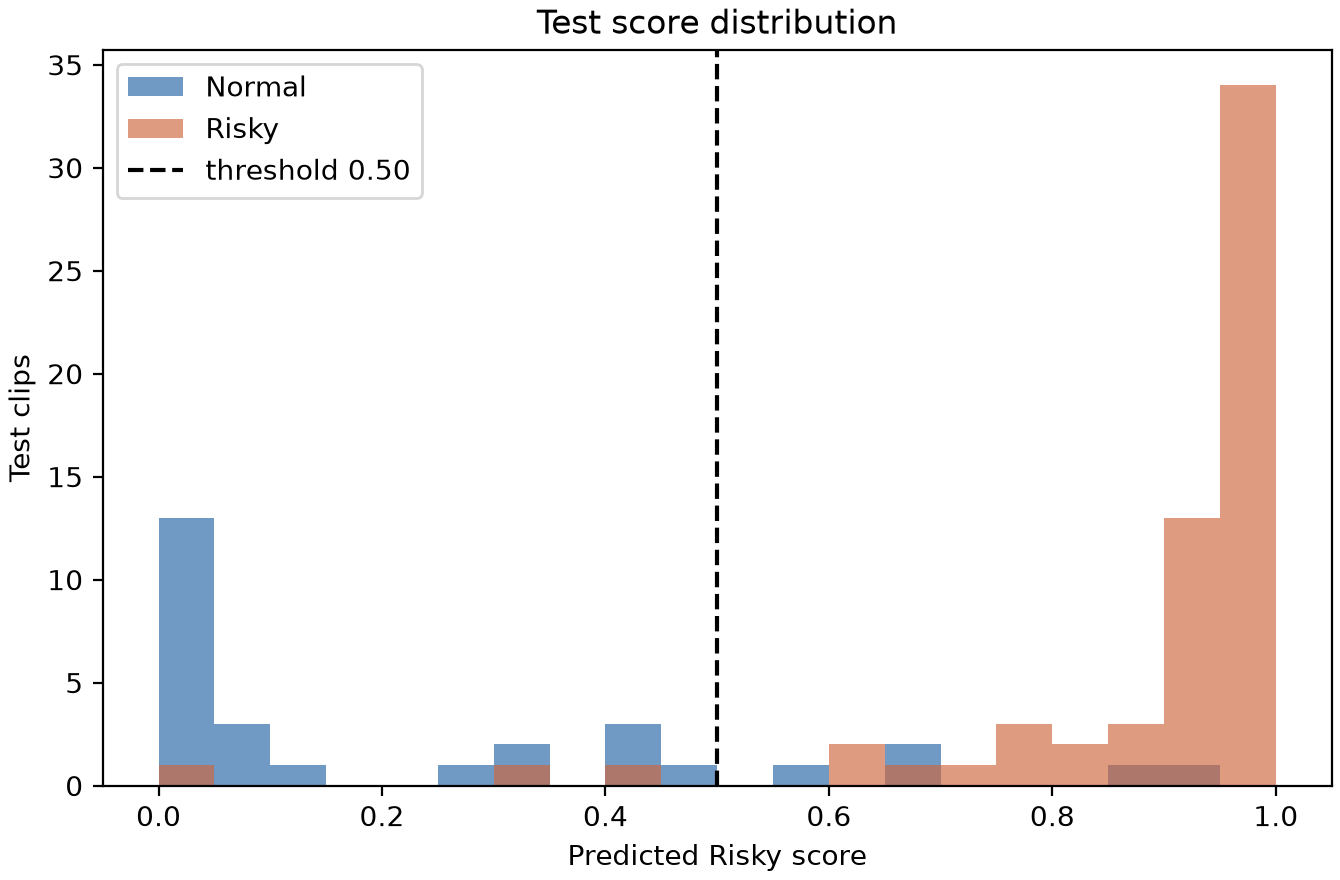}
    \caption{
    Distribution of predicted Risky probabilities for Normal and Risky clips
    in the held-out test set. Most Normal clips receive scores near zero,
    whereas most Risky clips receive scores near one, showing clear separation
    between the two classes. The dashed line indicates the fixed classification
    threshold of $0.5$. Blue observations to the right of the threshold
    correspond to false positives, while orange observations to the left
    correspond to false negatives.
    }
    \label{fig:test_score_distribution}
\end{figure*}

Figure~\ref{fig:test_score_distribution} shows the distribution of predicted
Risky probabilities produced by the final three-modality DeiT-Tiny model on
the held-out test set. Most Normal clips receive scores close to zero, whereas
most Risky clips receive scores close to one, showing a clear separation
between the two classes. The dashed line represents the fixed decision
threshold of $0.5$.

The binary prediction is obtained as

\begin{equation}
\hat{y} =
\begin{cases}
\text{Normal}, & p(\text{Risky}\mid\mathcal{X}) < 0.5,\\
\text{Risky}, & p(\text{Risky}\mid\mathcal{X}) \geq 0.5,
\end{cases}
\end{equation}

where $\mathcal{X}$ represents the multimodal input. Normal samples appearing
to the right of the threshold correspond to false positives, while Risky
samples appearing to the left correspond to false negatives. The limited
overlap between the two distributions indicates that the model separates most
test clips with predicted scores away from the decision boundary, although a
small number of difficult or confidently incorrect cases remain.

Table~\ref{tab:risk_rubric} summarizes the annotation criteria used to assign
the four original risk levels. The labels are based on observable cues across
visual appearance, body motion, and audio. No-Risk clips contain stable and
non-aggressive behavior, while Low- and Medium-Risk clips represent
progressively stronger behavioral changes and escalation. High-Risk clips
contain stronger pre-incident cues occurring close to the annotated incident
onset. For the binary experiments, Low-, Medium-, and High-Risk clips are
grouped into the Risky class, while No-Risk clips form the Normal class.

\begin{table}[!htbp]
\centering
\caption{Pre-Incident Risk Annotation Rubric}
\label{tab:risk_rubric}
\small
\setlength{\tabcolsep}{4pt}
\renewcommand{\arraystretch}{1.15}

\begin{tabularx}{\textwidth}{
>{\centering\arraybackslash}p{1.5cm}
>{\justifying\arraybackslash}X
>{\justifying\arraybackslash}X
>{\justifying\arraybackslash}X
>{\justifying\arraybackslash}X}

\hline
\textbf{Label} &
\textbf{Vision} &
\textbf{Motion} &
\textbf{Audio} &
\textbf{Risk Interpretation} \\
\hline

\multirow[c]{1}{1.5cm}{\centering\textbf{No Risk}} &
Individuals appear calm and non-aggressive. No threatening posture, weapon visibility, or confrontation is observed. People may be walking, standing, sitting, or interacting normally. &
Minimal body movement and stable trajectories. Normal interpersonal distance with smooth and predictable motion patterns. No sudden acceleration or aggressive gestures. &
Background ambience or normal conversation only. No shouting, screaming, impact sounds, or sudden loud audio spikes. &
The environment appears safe and stable with no visible indicators of aggression or escalation. \\
\hline

\textbf{Low Risk} &
Mild suspicious or tense visual behavior may appear, such as prolonged staring, following behavior, defensive posture, or unusual attention toward another individual. No direct aggression occurs. &
Slight increase in movement intensity or interpersonal focus. Gradual approach between individuals, mild trajectory changes, or increased body orientation toward another person. &
Slightly elevated audio activity such as louder speech, crowd noise, or environmental tension without explicit aggressive sounds. &
Early behavioral indicators suggest potential tension or discomfort, but no immediate threat is observed. \\
\hline

\textbf{Medium Risk} &
Clear behavioral escalation becomes visible. Aggressive body posture, arm raising, threatening gestures, chasing behavior, or visible confrontation may appear. &
Noticeable increase in motion intensity, abrupt trajectory changes, reduced interpersonal distance, and rapid body movements indicating possible conflict escalation. &
Raised voices, arguments, aggressive speech tone, sudden audio spikes, or increased environmental disturbance may be present. &
Behavioral patterns indicate a significant likelihood of an aggressive event occurring soon. The situation is unstable and escalating. \\
\hline

\textbf{High Risk} &
Immediate pre-incident behavior is strongly visible. Direct confrontation, attack preparation, weapon handling, striking posture, or physically aggressive interaction is clearly observed before the incident starts. &
Highly erratic and rapid motion patterns, aggressive acceleration, physical engagement attempts, collision-like movement, or intense approach dynamics are observed. &
Screaming, impact sounds, weapon-related sounds, explosions, or highly elevated audio intensity may occur immediately before the event. &
The scene shows strong pre-incident aggression cues indicating an imminent violent or dangerous event. \\
\hline

\end{tabularx}
\end{table}

\end{document}